\documentclass[conference]{IEEEtran}
\IEEEoverridecommandlockouts

\usepackage{cite}
\usepackage{amsmath,amssymb,amsfonts}
\usepackage{algorithmic}
\usepackage{algorithm}
\usepackage{graphicx}
\usepackage{textcomp}
\usepackage{xcolor}
\usepackage{booktabs}
\usepackage{multirow}
\usepackage{enumitem}
\usepackage{xcolor}
\usepackage{graphicx}
\usepackage{subcaption}
\usepackage{hyperref}
\def\BibTeX{{\rm B\kern-.05em{\sc i\kern-.025em b}\kern-.08em
    T\kern-.1667em\lower.7ex\hbox{E}\kern-.125emX}}

\begin{document}

\title{Geometric Coherence via Weighted Matching for\\
3D Heterogeneous Multi-Agent Reach-Avoid Games}

\author{Prajwal Vijay%
  \thanks{P. Vijay is with the Dept. of Electrical Engineering, 
  IIT Madras, Chennai, India. 
  {\tt ee23b057@smail.iitm.ac.in}}%
}

\maketitle

\begin{abstract}
We study assignment quality in 3D heterogeneous multi-agent reach-avoid games and identify a recurring failure mode of cardinality-only matching in geometrically structured scenarios, which we term \emph{Geometric Sprawl}. In these cases, multiple maximum-cardinality assignments are available, but some induce spatially incoherent pairings and inefficient pursuit trajectories. Building on the evasion-space framework of Yan et al.~\cite{yan2022}, we introduce a cardinality-first weighted sequential matching method in which the Hamilton--Jacobi--Isaacs interception value $z_I(s,j)$ is used as a secondary assignment weight. Each sequential stage is solved with a min-cost max-flow backend, while the unweighted baseline uses the same solver with the weight term removed. We evaluate both methods on a deterministic 35-scenario benchmark (7 families × 5 initialization variants) under two regimes: a diagnostic stationary-unmatched setting and a hybrid saddle-point setting with goal-directed unmatched evaders. On this benchmark, the weighted method resolves 13 of 15 geometric stress-test instances that cause repeated timeouts for the unweighted baseline in the diagnostic regime, and under the hybrid regime improves mean captures from $3.20$ to $3.91$ (22\% increase) and mean interception height from $3.87$ to $5.36$ (40\% increase). We also report lower path tortuosity and lower angular-effort proxy values, suggesting smoother pursuit trajectories in this first-order simulation model. We release the simulator and benchmark suite at \href{https://github.com/Prajwal-Vijay/geometric-coherence-weighted-matching}{github.com/Prajwal-Vijay/geometric-coherence-weighted-matching} to support reproducible evaluation of assignment strategies for 3D reach-avoid games.

\end{abstract}

\section{Introduction}

\subsection{The Resolution Gap}
Reach-avoid games model scenarios in which a team of pursuers attempts to intercept a set of evaders before they reach a goal region. Such problems arise naturally in perimeter defense, airspace protection, and autonomous security systems. We study the multi-agent setting in 3D, where a small defense team must coordinate to intercept a numerically superior group of intruders.
Consider three UAV defenders guarding a perimeter against ten intruders
in 3D space, where each evader seeks to reach the ground plane ($z=0$)
and each pursuer must intercept them before breach.
The pioneering work of Yan et al.~\cite{yan2022} reduces this to a
bipartite matching problem: enumerate all feasible pursuer-coalition--evader
pairs and maximize the number of captured evaders via network flow.
This is theoretically sound.  But it is physically fragile.

In each of these failure modes, the pursuer spends its motion budget on lateral or vertical repositioning rather than forward closure. The interception height decreases monotonically as the pursuer diverges from its geometric corridor. In our diagnostic study (where unmatched evaders remain stationary), this lateral oscillation prevents the engagement from ever resolving, causing state of the art (SOTA) unweighted baseline of Yan et al. ~\cite{yan2022} to hit the 2000-step timeout with 2–3 evaders still left on the board. Our benchmark (Table \ref{tab:static}) confirms this: the unweighted baseline times out on all 5 split-lanes variants, all 5 x-crossing variants, and 4/5 switchback-gate variants, while our weighted method resolves all of these.

We call this failure mode \textbf{Geometric Sprawl}: the tendency of
cardinality-only matching, in the presence of combinatorial degeneracy,
to select assignments that are feasible but spatially incoherent,
producing unnecessarily convoluted pursuit trajectories and mission
timeouts.

Our solution is to use the HJI value function---the $z$-height of the
interception point $z_{\mathcal{I}(s,j)}$---as an edge weight in the
bipartite assignment graph, and to solve the resulting lexicographic
two-objective problem via MCMF.  The weight acts as a geometric
regularizer: same-side, directionally coherent assignments produce higher interception heights (because the pursuer is already aligned with the
evader's corridor), while cross-arena assignments produce lower heights
(because lateral traversal delays closure).  The MCMF solver therefore
implicitly prefers spatial coherence without any explicit non-crossing
constraint.

\subsection{Contributions}

\begin{enumerate}[leftmargin=*,noitemsep]
  \item We empirically characterize \emph{\textbf{Geometric Sprawl}}, a failure mode of cardinality-only matching in spatially structured 3D reach-avoid scenarios.
  \item \textbf{Weighted MCMF formulation} using the HJI value function
        as a geometric regularizer, resolving Geometric Sprawl while
        preserving the $\frac{1}{3}$-approximation cardinality guarantee
  \item \textbf{Four geometric stress-test families} (Split Lanes, X-Crossing,
        Braided Corridors, Switchback Gate) designed to expose the unweighted baseline's
        failure modes and benchmark weighted assignment.
  \item Evaluation under a \textbf{numerically overwhelming hybrid saddle-point regime}, demonstrating a 22\% cardinality improvement and a 39\% interception quality gain over the unweighted baseline.
  \item \textbf{Physical feasibility metrics} (path tortuosity $\tau$,
        angular control effort $J_\theta$, fleet-wide energy) showing
        that weighted matching produces robotically realistic trajectories that cardinality-only matching fails to produce in geometrically structured scenarios.
  \item We release a \textbf{modular python simulation framework} and deterministic benchmark suite to support future research in geometry-aware multi-agent assignment.
\end{enumerate}

\section{Related Work}

\subsection{Reach-Avoid Differential Games}

The study of pursuit-evasion differential games originates with
Isaacs~\cite{isaacs1965}, whose HJI framework characterizes game-theoretic
value functions.  Hamilton-Jacobi formulations for two-player reach-avoid
games were developed by Margellos and Lygeros~\cite{margellos2011} and
Mitchell et al.~\cite{mitchell2005}, but suffer from the curse of
dimensionality.  Chen et al.~\cite{chen2017} addressed multiplayer 2D
settings via pairwise outcomes and unweighted network flow.  Yan
et al.~\cite{yan2020} extended to convex planar domains via analytical
barriers.  In~\cite{yan2022}, the same authors generalized to 3D,
heterogeneous speeds, and non-zero capture radii, proving that coalitions
of at most three pursuers suffice and formulating the assignment as a
Sequential Matching Algorithm with unweighted network flow.

Alternative approaches to multiplayer reach-avoid scenarios include mixed-integer programming formulations \cite{lorenzetti2018}, real-time active defense heuristics \cite{mohanan2019} and analytical differential game solutions \cite{garcia2019}; however, these generally struggle to scale dynamically or lack the strict geometric closure guarantees provided by the Evasion Space framework.

\textit{The critical gap:} in all of the above works, the matching
objective is pure cardinality.  No prior work has incorporated the HJI
value function as an assignment weight, nor used MCMF in this context.

\subsection{MCMF in Multi-Agent Systems}

Maximum-cost maximum-flow algorithms are well-established for multi-robot
task assignment~\cite{ford1962,ahuja1993}.  However, their application
to reach-avoid games with game-theoretic edge weights is novel to this
work.

\subsection{Simulation Platforms}

Existing simulation tools for pursuit-evasion are largely restricted to
2D environments, homogeneous agents, or learning-based frameworks lacking
formal guarantees.  No publicly available platform exists for the 3D
heterogeneous reach-avoid game with coalition-based ES strategies.

\section{Problem Formulation}

\subsection{Multiplayer Reach-Avoid Differential Game}

Consider a reach-avoid game with $N_p$ pursuers
$\mathcal{P}=\{P_1,\ldots,P_{N_p}\}$ and $N_e$ evaders
$\mathbb{E}=\{E_1,\ldots,E_{N_e}\}$ in $\mathbb{R}^3$.
The goal region is $\Omega_{\text{goal}}=\{x\in\mathbb{R}^3\mid z\le 0\}$
and the play region is $\Omega_{\text{play}}=\{x\in\mathbb{R}^3\mid z>0\}$.

All players are holonomic mass points with first-order
kinematics~\cite{yan2022}:
\begin{equation}
\dot{x}_{P_i}(t)=v_{P_i}u_{P_i}(t),\quad
\dot{x}_{E_j}(t)=v_{E_j}u_{E_j}(t),
\label{eq:dynamics}
\end{equation}
where $u_{P_i}(t),u_{E_j}(t)\in\mathbb{S}^2$ are heading unit vectors
and $v_{P_i},v_{E_j}\in\mathbb{R}^+$ are speeds.  Pursuer $P_i$ has
capture radius $r_i\ge 0$; evader $E_j$ is captured when
$\|x_{E_j}-x_{P_i}\|_2\le r_i$.

\textit{Assumptions} (following~\cite{yan2022}): \textbf{(A1)} all players
begin at distinct positions; no evader initially within any capture
radius; \textbf{(A2)} speed ratio $\alpha_{ij}=v_{P_i}/v_{E_j}>1$ for
all $(P_i,E_j)$; \textbf{(A3)} state-feedback control.

\section{Theoretical Background} 

\subsection{Evasion Space and Potential Functions}

For a coalition $\mathcal{P}_s=\{P_i\mid i\in s\}$ and evader $E_j$,
the Evasion Space $\mathcal{E}(s,j)$ is the set of positions $E_j$ can
reach without capture, characterized by~\cite{yan2022}:
\begin{align}
f_{ij}(x)&=\|x-x_{P_i}\|_2-\alpha_{ij}\|x-x_{E_j}\|_2-r_i,
\label{eq:potential}\\
\bar{\mathcal{E}}(s,j)&=\bigl\{x\in\mathbb{R}^3\bigm|
f_{ij}(x)\ge 0,\;\forall\,i\in s\bigr\}.
\label{eq:ES}
\end{align}
For $r_i=0$, $\partial\mathcal{E}(i,j)$ reduces to the classical
Apollonius sphere; for $r_i>0$ it is a strictly convex generalization.
The set $\bar{\mathcal{E}}(s,j)$ is bounded and strictly
convex~\cite{yan2022}.

\subsection{Interception Point and HJI Value Function}

When $\bar{\mathcal{E}}(s,j)\cap\Omega_{\text{goal}}=\emptyset$,
coalition $\mathcal{P}_s$ can prevent $E_j$ from reaching the goal.
The interception point $\mathcal{I}(s,j)$ is:
\begin{equation}
\mathcal{I}(s,j)=\arg\min_{x\in\mathbb{R}^3}z
\quad\text{s.t.}\quad f_{ij}(x)\ge 0,\;\forall\,i\in s.
\label{eq:intercept}
\end{equation}
Each pursuer in $\mathcal{P}_s$ moves toward $\mathcal{I}(s,j)$; this
guarantees $\dot{z}_{\mathcal{I}(s,j)}\ge 0$ for any evader
control. The optimal control for the evader in this case is to move toward $\mathcal{I}(s,j)$, which ensures that $\dot{z}_{\mathcal{I}(s,j)}= 0$ ~\cite{yan2022}. The HJI value function equals the $z$-coordinate
of the optimal interception point:
\begin{equation}
V(x_s,x_{E_j})=z_{\mathcal{I}(s,j)}.
\label{eq:VF}
\end{equation}
A larger $V$ means the evader is intercepted deeper in the play
region---a strategically superior outcome for the defense.

\subsection{At Most Three Pursuers Suffice}

A fundamental result from~\cite{yan2022}: for any coalition
$\mathcal{P}_s$ capable of capturing $E_j$, there exists a sub-coalition
$s_1\subseteq s$ with $|s_1|\le 3$ such that
$\mathcal{I}(s_1,j)=\mathcal{I}(s,j)$.
This bounds the coalition enumeration and keeps the matching tractable.

\section{Weighted Matching: Resolving Geometric Sprawl}

\subsection{Why Cardinality-Only Matching Fails in Structured Geometries}

The SOTA formulation~\cite{yan2022} maximizes only the number of captured
evaders.  A coalition intercepting an evader at $z=0.01$ above the goal
plane is treated identically to one catching an evader at $z=7.0$---both
count as one capture.  This indifference to capture quality is acceptable
when the assignment has a unique maximum-cardinality matching.  It is
catastrophic when many such matchings exist.

In geometrically structured scenarios---evaders in separated lanes,
crossing corridors, or staggered gates---many distinct assignments achieve
the same maximum cardinality.  The cardinality-only solver selects among
these by network topology or index ordering, which is uncorrelated with spatial geometry.  This can produce:
\begin{itemize}[noitemsep,leftmargin=*]
  \item \textit{Anti-diagonal assignments:} a left-side pursuer assigned
        to a right-side evader, and vice versa.  Both are feasible;
        combined, they require each pursuer to traverse the full arena
        width before converging.
  \item \textit{Corridor-switching:} a pursuer oscillating between
        evaders in different spatial lanes because both are equally
        ``cardinality-optimal.''  The pursuer accumulates lateral
        displacement instead of vertical closure.
  \item \textit{Gate-alternation:} a pursuer assigned to alternate
        between lower and upper defensive gates instead of clearing one gate locally, producing repeated heading reversals.
\end{itemize}

In all three cases, the pursuer spends its motion budget on lateral or vertical repositioning rather than forward closure. The interception height decreases monotonically as the pursuer diverges from its geometric corridor, until Evasion Space closure is permanently lost. The remaining evaders become mathematically uncatchable, resulting in empty assignments. In our diagnostic study (where unmatched evaders remain stationary), these broken engagements simply run out the clock, causing the unweighted baseline to hit the 2000-step timeout with 2–3 evaders still left on the board.

\subsection{The Weighted Bipartite Graph}
We construct a weighted bipartite graph $G=(U\cup V,\mathcal{E}_G,w)$:
\begin{itemize}[noitemsep,leftmargin=*]
  \item $U=[P]^{\le 3}$: all pursuit coalitions of size $\le 3$.
  \item $V=\mathcal{E}$: the evader set.
  \item $e_{sj}\in\mathcal{E}_G$ iff $\mathcal{P}_s$ can capture $E_j$
        but no proper sub-coalition can.
  \item Edge weight: $w_{sj}=V(x_s,x_{E_j})=z_{\mathcal{I}(s,j)}$.
\end{itemize}
Higher $w_{sj}$ corresponds to a deeper interception.  The conflict graph
$\mathcal{C}=(\mathcal{E}_G,\bar{E})$ encodes pairs of edges that share a
pursuer and cannot coexist.

\subsection{MBMC Formulation and Lexicographic Objective}

We formulate the weighted assignment as a Maximum Bipartite Matching with Cost (MBMC) problem:
\begin{align}
\text{maximize}&\;\sum_{e_{sj}\in\mathcal{E}_G}a_{sj}
  &&\tag{primary: cardinality}\label{eq:obj1}\\
\text{then maximize}&\;\sum_{e_{sj}\in\mathcal{E}_G}w_{sj}a_{sj}
  &&\tag{secondary: quality}\label{eq:obj2}\\
\text{s.t.}&\;\sum_s a_{sj}\le 1\;\;\forall E_j,\label{eq:c1}\\
           &\;\sum_j a_{sj}\le 1\;\;\forall\mathcal{P}_s,\label{eq:c2}\\
           &\;a_{sj}+a_{pq}\le 1\;\;
             \forall(e_{sj},e_{pq})\in\bar{E},\label{eq:c3}\\
           &\;a_{sj}\in\{0,1\}.\label{eq:c4}
\end{align}
This is a strict generalization of~\cite{yan2022}, which optimizes the
primary objective alone.  Among all maximum-cardinality matchings, we
select the one with greatest total interception height.  Quality never
degrades cardinality.

\subsection{MCMF via Successive Shortest Paths with Johnson's Reweighting}

The MBMC is solved via the three-stage Sequential Matching
Algorithm~(Algorithm~\ref{alg:seq}), with the unweighted flow
subproblem at each stage replaced by a min-cost max-flow (MCMF)
problem. The MCMF is solved using the Successive Shortest Path (SSP) 
algorithm with vertex potentials (Johnson reweighting) \cite{ahuja1993} to maintain
non-negative reduced costs, thereby allowing the use of Dijkstra's
algorithm instead of Bellman--Ford. The worst-case complexity of one
MCMF solve is $O(|E|\,|V|\log |V|)$.

\begin{algorithm}[ht]
\caption{Weighted Sequential Matching ~\cite{yan2022}}
\label{alg:seq}
\begin{algorithmic}[1]
\REQUIRE Weighted bipartite graph $G=(U\cup V,\mathcal{E}_G,w)$
\ENSURE Approximate weighted matching $M$
\STATE $U_1\leftarrow\mathcal{P}$, $V_1\leftarrow\mathcal{E}$
\STATE $M_1\leftarrow\text{MCMF}(G_1)$ \COMMENT{1-vs-1 stage}
\STATE $A_1\leftarrow$ matched pursuers; $B_1\leftarrow$ matched evaders
\STATE $U_2\leftarrow[\mathcal{P}\setminus A_1]^2$,
       $V_2\leftarrow\mathcal{E}\setminus B_1$
\STATE $M_2\leftarrow\text{MCMF}(G_2)$ \COMMENT{2-vs-1 stage}
\STATE $A_2\leftarrow$ matched pursuers; $B_2\leftarrow$ matched evaders
\STATE $U_3\leftarrow[\mathcal{P}\setminus(A_1\cup A_2)]^3$,
       $V_3\leftarrow\mathcal{E}\setminus(B_1\cup B_2)$
\STATE $M_3\leftarrow\text{MCMF}(G_3)$ \COMMENT{3-vs-1 stage}
\RETURN $M=M_1\cup M_2\cup M_3$
\end{algorithmic}
\vspace{0em}
\end{algorithm}

\textit{Encoding the lexicographic objective.}
Edge cost of $e_{sj}$ is set to $-(M_{\text{large}}+w_{sj})$, where
$M_{\text{large}}>\sum_{e\in\mathcal{E}_G}w_{sj}$.  This guarantees any
matching with one additional capture is always preferred over any
matching with higher total weight but fewer captures.  The
$\frac{1}{3}$-approximation cardinality guarantee of~\cite{yan2022} is
preserved. Since $M_{\text{large}}$ dominates any weight sum, the MCMF solver prioritizes an additional capture over any quality gain, leaving the cardinality-maximizing stage of the SSP algorithm structurally identical to the unweighted case.


\section{Simulation Framework and Benchmark Suite}

\subsection{Framework Architecture}

\begin{figure}
    \centering
    \includegraphics[width=1\linewidth]{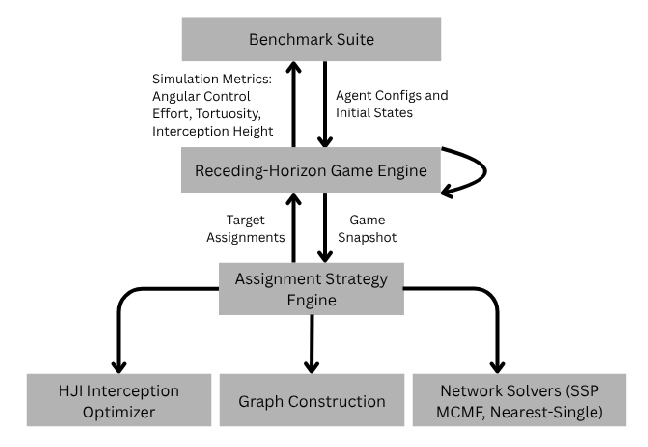}
    \caption{Architecture Diagram}
    \label{fig:architecture}
\end{figure}

The simulation is implemented in Python, utilizing SciPy for sequential least squares programming (SLSQP) and Matplotlib for 3D visualization. To support reproducible ablation studies, the architecture is structured into three interacting tiers as depicted in figure \ref{fig:architecture}:

\textbf{Scenario \& Evaluation Tier}: Generates the 35-scenario deterministic benchmark suite. It initializes heterogeneous agent configurations (speeds, capture radii) and logs fleet-wide telemetry, including path tortuosity and angular control effort ($J_{\theta}$).

\textbf{Receding-Horizon Simulation Tier}: The core discrete-time game engine ($\Delta t=0.1$ s). It manages agent kinematics, continuously evaluates dynamic capture/escape conditions via distance thresholds and goal-plane boundaries, and triggers event-driven replanning when the game state changes.

\textbf{Algorithmic Backend Tier}: Houses the decision intelligence. It consists of a convex optimization module to compute the HJI value functions, a graph constructor that builds the weighted bipartite structures and conflict graphs, and an abstracted assignment interface that routes the graph to the chosen solver (e.g., the Successive Shortest Path MCMF engine or greedy heuristics).

\subsection{The 35-Scenario Deterministic Benchmark Suite}

Each family has five positional and speed-perturbation variants (v1–v5), yielding 35 distinct scenarios. The five variants serve as a robustness check: by averaging metrics across perturbed initializations, we ensure reported performance differences reflect assignment strategy rather than sensitivity to a single configuration.

All families use a 3-pursuer vs.\ 10-evader configuration with
heterogeneous speeds ($v_P\in[1.45,1.82]$~m/s,
$v_E\in[0.90,1.16]$~m/s) and capture radius $r_i=0.1$.


The three stress-test families are specifically engineered to expose
Geometric Sprawl:

\textbf{Split Lanes.} Three pursuers in an inverted-triangle formation guard evaders segregated into two parallel horizontal corridors (an upper and a lower lane). The efficient decomposition allocates pursuers strictly to their respective local lanes. Because cardinality-only solvers are indifferent to cross-lane displacement, they frequently assign pursuers to opposite lanes, inducing weaving trajectories and high path tortuosity. The weighted objective heavily penalizes this lateral transit, enforcing strict lane discipline by naturally selecting same-corridor assignments that yield higher interception heights.

\textbf{X-Crossing.}  Two low-lateral pursuers and one upper-central
pursuer guard ten evaders in two opposing lateral packets.  The efficient
decomposition assigns same-side pursuers to same-side evaders.
Cardinality-only matching can produce anti-diagonal assignments that force
arena-crossing trajectories.  The weighted objective breaks this
degeneracy because same-side assignments yield higher $z_\mathcal{I}$:
the pursuer is already aligned with the evader's corridor of motion.

\textbf{Braided Corridors.}  Two longitudinal corridors (upper and lower
band) with evader indices deliberately interleaved, scrambling
lexicographic tie-breaking.  The efficient assignment preserves corridor
membership; cardinality-only matching with index-based tie-breaking
alternates between corridors, producing weaving paths.  The weighted
objective selects corridor-consistent assignments because they carry
substantially higher $z_\mathcal{I}$.

\textbf{Switchback Gate.}  Two staggered defensive gates with side
pursuers shifted inward.  The efficient response clears nearby lower-gate
evaders first, then folds upward---a monotone two-phase sweep.
Cardinality-only matching can alternate a pursuer between lower and upper
gates, producing repeated heading reversals.  The weighted objective
suppresses this because monotone gate progress yields consistently higher
interception heights than oscillatory gate-switching.

\subsection{Evader Policy and Experimental Design}

\textbf{Diagnostic study (static benchmark).}  Unmatched evaders remain
stationary until a pursuit coalition is assigned; once matched, each
evader moves toward the computed interception point $\mathcal{I}(s,j)$
(the saddle-point optimal strategy from~\cite{yan2022}).  This regime
isolates the assignment logic from adversarial motion.  Mission delays
and timeouts are strictly attributable to Geometric Sprawl in the
matching, not to evader maneuvering.

\textbf{Adversarial stress test (Hybrid Saddle-Point Policy).} In this regime, unassigned evaders execute a greedy breach strategy, heading toward $z=0$ at maximum speed. Meanwhile, matched evaders employ the theoretically optimal HJI subgame strategy, moving actively toward the interception point $\mathcal{I}(s,j)$, i.e., the saddle-point optimal strategy, which minimizes interception height for the evader while the pursuit team maximizes it. This is the most challenging regime: a numerically overwhelming scenario that tests whether the pursuit team can extract maximum value from the engagements it can win

\section{Results}
\textit{Performance metrics}. We report steps to resolution and termination reason (\texttt{all\_resolved} vs.\ \texttt{max\_steps}), captured/escaped evader counts, mean interception height
\[
\bar z_I = \frac{1}{|\mathrm{cap}|}\sum z_I(s,j),
\]
and two trajectory-quality metrics: path tortuosity $\tau = L_{\mathrm{path}}/L_{\mathrm{chord}}$ and angular-effort proxy
\[
J_\theta = \sum_t \|\Delta \hat u_t\|_2^2 / \Delta t.
\]
In our first-order holonomic simulator, lower $\tau$ and $J_\theta$ indicate smoother and less aggressively turning trajectories. We therefore interpret these as simulation-level proxies for trajectory complexity and control effort, rather than as direct measurements of battery life, actuator wear, or flight-envelope limits.

\begin{figure*}[htbp]
     \centering
     \begin{subfigure}[b]{0.45\textwidth}
         \centering
         \includegraphics[width=\textwidth]{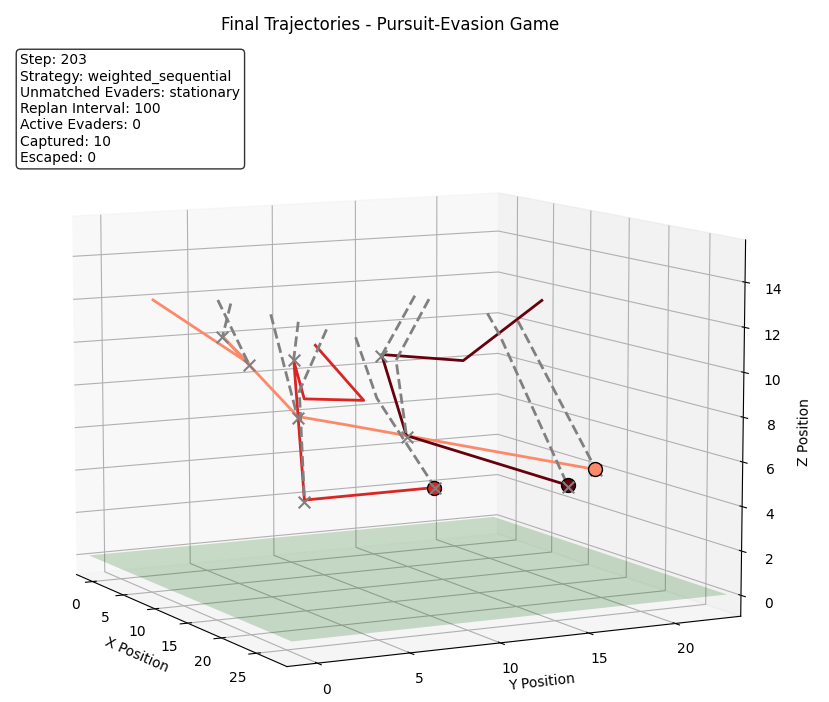}
         \caption{\textbf{Braided Corridors weighted} The Pursuers sequentially capture the evaders without going back and forth between the corridors and maintain clear responsibility for the rougly 'parallel' corridors.}
         \label{fig:image1}
     \end{subfigure}
     \hfill
     \begin{subfigure}[b]{0.45\textwidth}
         \centering
         \includegraphics[width=\textwidth]{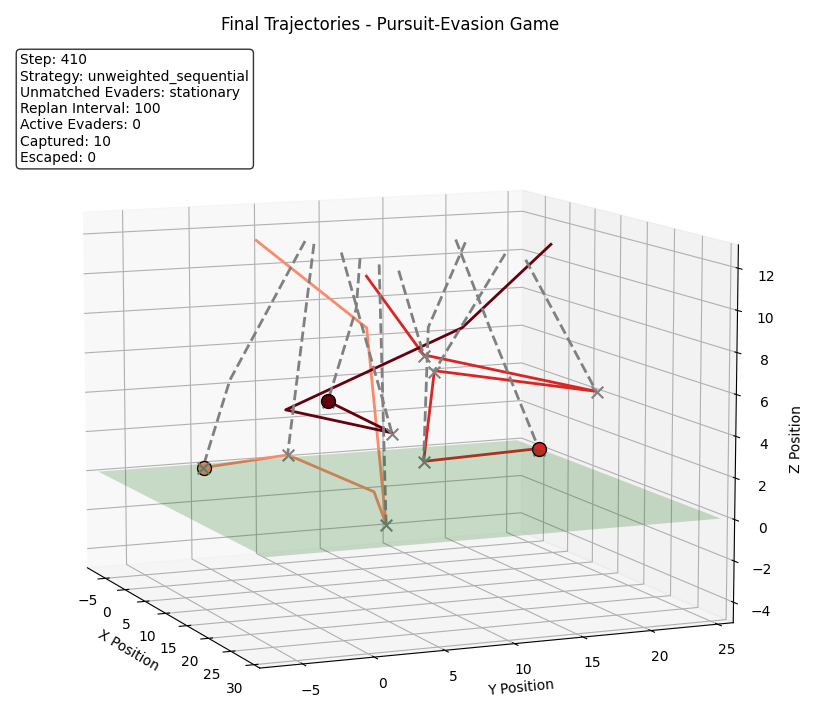}
         \caption{\textbf{Braided Corridors Unweighted} Matching blindness leads to geometric sprawl. This unweighted approach leads to chaotic criss-crossing of assignments, where a pursuer on the left(orange) is winding sharply right to address the rightmost evader.}
         \label{fig:image2}
     \end{subfigure}

     \vspace{10pt} 

     \begin{subfigure}[b]{0.45\textwidth}
         \centering
         \includegraphics[width=\textwidth]{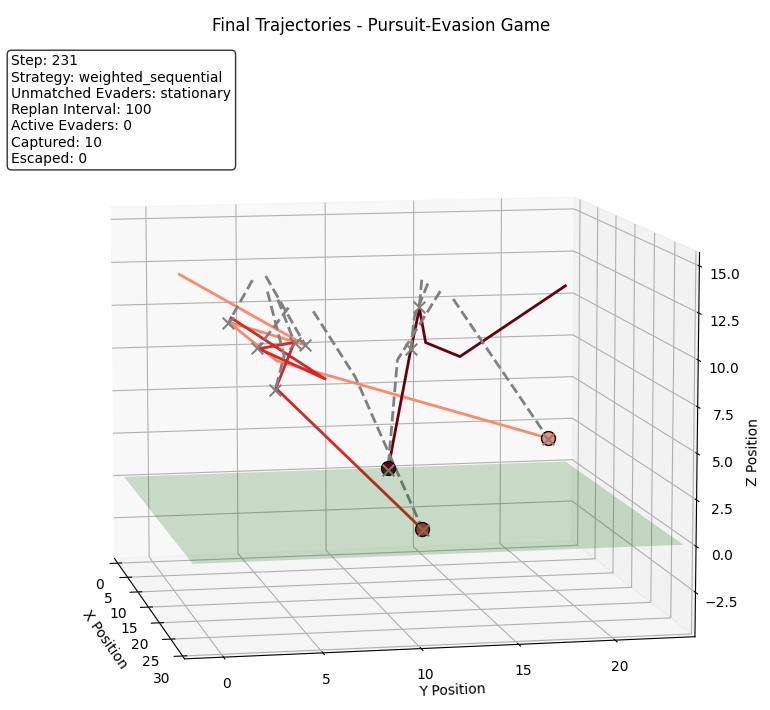}
         \caption{\textbf{Switchback Gate weighted} Shows orderly and spatially coherent clearing in the switchback scenario. Assignments are local and efficient, pursuers sweep their respective evaders in mostly monotone, local trajectories.}
         \label{fig:image3}
     \end{subfigure}
     \hfill
     \begin{subfigure}[b]{0.45\textwidth}
         \centering
         \includegraphics[width=\textwidth]{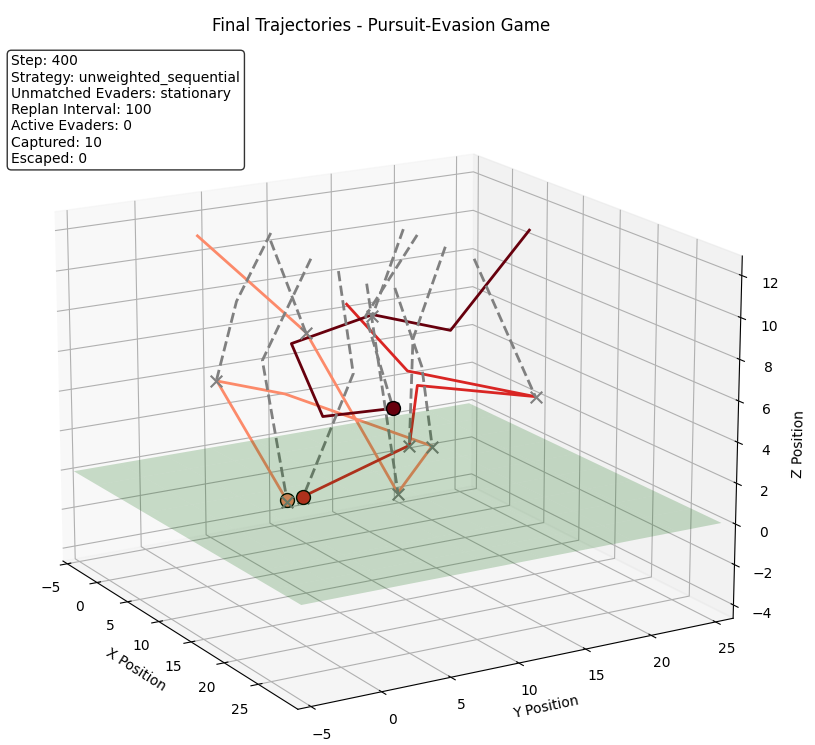}
         \caption{\textbf{Switchback Gate Unweighted} Displays severe geometric sprawl and path chaos in the same scenario. Unweighted matching creates inefficient initial pairings, forcing pursuers into highly winding, highly non-monotone paths}
         \label{fig:image4}
     \end{subfigure}
        
     \caption{Trajectory comparison between weighted (left, (a)(c)) and unweighted (right, (b)(d)) sequential matching under the diagnostic stationary-evader regime. Braided Corridors (top, (a)(b)) and Switchback Gate (bottom, (c)(d)) families shown.}
     \label{fig:four_graphs}
\end{figure*}

\subsection{Diagnostic Study: Isolating Geometric Sprawl}

Table~\ref{tab:static} presents per-family averages under the static
benchmark.  Two regimes emerge clearly.

\textbf{Baseline families} (spread-mild-hetero, clustered-center,
asymmetric-fast) show that both weighted and unweighted strategies
resolve comparable numbers of evaders within the step budget, confirming
that the two methods are equivalent in the absence of geometric
structure.  The weighted method consistently achieves higher
$\bar{z}_\mathcal{I}$ and lower $\bar{\tau}$, but neither method
triggers systematic timeouts.

\textbf{Geometric stress tests} reveal the failure mode sharply.
The unweighted baseline times out on \emph{all 5} x-crossing variants,
\emph{all 5} split-lanes variants, and 4/5 switchback-gate variants.
Our weighted method resolves 5/5 split-lanes, 4/5 x-crossing, and 5/5
switchback-gate---13 of 15 stress-test instances compared to 1 of 15 for the SOTA.

The tortuosity data in the switchback-gate family provides the clearest
physical evidence of Geometric Sprawl.  The unweighted baseline achieves
$\bar{\tau}=5.17$: the average pursuer travels five times the
straight-line distance to its final position.  A tortuosity of 5.17
means the path doubles back on itself repeatedly---gate-alternation
producing a switchback trail.  Our method achieves $\bar{\tau}=1.82$,
indicating near-monotone local clearing.  This 184\% difference in tortuosity represents more than just spatial inefficiency; while our evaluation utilizes first-order holonomic dynamics, this significant reduction suggests that the resulting trajectories are inherently more compatible with the strict bounded angular rates and dynamic constraints of real-world UAVs.

The fleet-wide energy argument is equally stark in split-lanes.
Our method resolves all five variants at an average of 268~steps with
$\bar{\tau}=2.00$.  The unweighted baseline runs for the full 2000~steps on all five
variants.

\begin{table}[t]
\centering
\caption{Diagnostic study (stationary-unmatched evader policy):
per-family averages over 5 variants, 3 pursuers vs.\ 10 evaders,
\texttt{max\_steps}$=2000$.
$\bar{z}_\mathcal{I}$: avg.\ interception height (higher is better);
$\bar{\tau}$: avg.\ path tortuosity (lower is better);
$J_\theta$: avg.\ angular control effort (rad);
For sequential strategies, Esc.$=0$ at \texttt{max\_steps} means
remaining evaders became mathematically unresolvable (empty assignments) and timed out.}
\label{tab:static}
\setlength{\tabcolsep}{3pt}
\begin{tabular}{llcccccc}
\toprule
\textbf{Family} & \textbf{Strategy} &
\textbf{Steps} & \textbf{Capt.} & \textbf{Esc.} &
$\bar{z}_\mathcal{I}$ & $\bar{\tau}$ & $J_\theta$ \\
\midrule
\multirow{4}{*}{spread-mild}
 & Weighted (ours)   & 2000 & 8.2 & 0.0 & \textbf{7.21} & \textbf{1.78} & 108 \\
 & Unweighted~\cite{yan2022} & 2000 & 8.2 & 0.0 & 3.55 & 3.64 & 259\\
 & Random            & 2000 & 7.2 & 0.0 & 4.69 & 3.14 & 221 \\
 & Nearest-single    & 305  &\textbf{10.0}& 0.0 & 6.09 & 3.03 & 252 \\
\midrule
\multirow{4}{*}{clustered}
 & Weighted (ours)   & 2000 & 8.2 & 0.0 & \textbf{8.58} & \textbf{2.73} & 259 \\
 & Unweighted~\cite{yan2022} & 1381 & 9.4 & 0.0 & 5.18 & 2.84 & 383 \\
 & Random            & 2000 & 7.8 & 0.0 & 4.85 & 3.43 & 362 \\
 & Nearest-single    & 298  & 8.8 & 1.2 & 6.12 & 3.16 & 295 \\
\midrule
\multirow{4}{*}{asym.-fast}
 & Weighted (ours)   & 2000 &\textbf{6.2}& 0.0 & \textbf{6.97} & \textbf{1.95} & 197 \\
 & Unweighted~\cite{yan2022} & 2000 & 3.0 & 0.0 & 4.43 & 1.51 & 66 \\
 & Random            & 2000 & 6.0 & 0.0 & 5.13 & 2.65 & 219 \\
 & Nearest-single    & 388  & 7.8 & 2.2 & 4.54 & 2.67 & 256 \\
\midrule
\multirow{4}{*}{split-lanes$^\star$}
 & Weighted (ours)   & \textbf{268}  &\textbf{10.0}& 0.0 & 6.71 & 2.00 & 223 \\
 & Unweighted~\cite{yan2022} & 2000 & 8.2 & 0.0 & 5.86 & \textbf{1.89} & 156 \\
 & Random            & 2000 & 8.2 & 0.0 & 3.74 & 3.26 & 254 \\
 & Nearest-single    & 280  & 9.6 & 0.4 & 6.35 & 2.07 & 160 \\
\midrule
\multirow{4}{*}{x-crossing$^\star$}
 & Weighted (ours)   & \textbf{576}  & 9.8 & 0.0 & \textbf{7.45} & \textbf{2.09} & 220 \\
 & Unweighted~\cite{yan2022} & 2000 & 9.0 & 0.0 & 4.47 & 3.13 & 180 \\
 & Random            & 1681 & 7.8 & 0.0 & 4.50 & 2.88 & 251 \\
 & Nearest-single    & 187  &\textbf{10.0}& 0.0 & 8.14 & 1.91 & 184 \\
\midrule
\multirow{4}{*}{braided$^\star$}
 & Weighted (ours)   & \textbf{565}  & 9.8 & 0.0 & \textbf{7.12} & \textbf{2.20} & 164 \\
 & Unweighted~\cite{yan2022} & 1048 & 9.2 & 0.0 & 3.38 & 3.48 & 210 \\
 & Random            & 1676 & 8.4 & 0.0 & 5.47 & 2.73 & 269 \\
 & Nearest-single    & 237  &\textbf{10.0}& 0.0 & 7.41 & 2.06 & 189 \\
\midrule
\multirow{4}{*}{switchback-gate$^\star$}
 & Weighted (ours)   & \textbf{231}  &\textbf{10.0}& 0.0 & \textbf{7.09} & \textbf{1.82} & 204 \\
 & Unweighted~\cite{yan2022} & 1680 & 9.2 & 0.0 & 4.14 & 5.17 & 221 \\
 & Random            & 2000 & 8.6 & 0.0 & 4.13 & 2.86 & 257 \\
 & Nearest-single    & 298  & 9.4 & 0.6 & 6.42 & 2.05 & 213 \\
\midrule
\midrule
\multirow{4}{*}{\textbf{Overall (35 scenarios)}}
 & Weighted (ours)   & \textbf{1091} & \textbf{8.9} & 0.0 & \textbf{7.30} & \textbf{2.08} & 196 \\
 & Unweighted~\cite{yan2022} & 1730 & 8.0 & 0.0 & 4.43 & 3.09 & 211 \\
 & Random            & 1908 & 7.7 & 0.0 & 4.65 & 2.99 & 262 \\
 & Nearest-single    & 285  & 9.4 & 0.6 & 6.44 & 2.42 & 221 \\
\bottomrule
\multicolumn{7}{l}{\footnotesize $^\star$ Geometric stress-test family.}
\end{tabular}
\end{table}

\subsection{Adversarial Stress Test: The Hybrid Saddle-Point Regime}
Under this hybrid regime, the pursuit team faces a numerically overwhelming scenario where unassigned evaders execute a greedy breach toward the goal plane, while assigned evaders maneuver optimally toward interception.  Most will breach; the question is
which few the pursuit team can intercept, and how effectively.
This is the ``capture at the margins'' regime: the
assignment quality determines not merely \emph{how many} are caught
but \emph{how well} each capture is executed.

Table~\ref{tab:down} presents per-family averages.  The capture counts
are now much lower for all strategies (the pursuit team is numerically
overwhelmed), making the quality metrics $\bar{z}_\mathcal{I}$ and
$J_\theta$ the primary discriminators.

\textbf{Overall capture and interception quality.}
Weighted sequential captures an average of 3.91 evaders per scenario
versus 3.20 for the unweighted baseline---a \textbf{22\% increase in capture
cardinality} under adversarial pressure.  The mean interception height
improves from $\bar{z}_\mathcal{I}=3.87$ (baseline) to $5.36$ (ours):
a \textbf{39\% improvement in strategic safety margin}.  Each additional
meter of interception height represents a larger buffer against
sensor noise, model mismatch, and residual evader maneuvering.

\textbf{The asymmetric-fast family: quality under scarcity.}
When evaders are too fast for any single pursuer, only one evader can
be intercepted per run.  Here, cardinality is identical (1.0 for all
strategies), but $\bar{z}_\mathcal{I}$ diverges dramatically:
$8.94$ (weighted) versus $3.29$ (unweighted)---a $172\%$ improvement.
The SOTA occasionally selects a coalition that intercepts near the
goal plane (some variants return $z_\mathcal{I}\approx 0.1$, barely
above breach), while the weighted objective always selects the highest
available intercept.

\textbf{Stress tests under adversarial pressure.}
In x-crossing and braided-corridors, weighted sequential captures 6.6
and 6.0 evaders per scenario, compared to 5.0 and 4.0 for the unweighted baseline.
These families have geometrically structured evader layouts; even under the hybrid saddle-point regime, the assignment geometry matters. The weighted objective immediately selects same-side or corridor-consistent
assignments, achieving closure faster and at higher altitude before the
evaders reach the goal plane.

\begin{table}[t]
\centering
\caption{Adversarial Stress Test (Hybrid Saddle-Point Regime): Per-Family Averages, 3 Pursuers VS. 10 Evaders. Unmatched evaders charge toward $z=0$, matched evaders maneuver to interception. All Episodes terminate \texttt{ALL\_RESOLVED} (Evaders either captured OR escaped).}
\label{tab:down}
\setlength{\tabcolsep}{3.5pt}
\begin{tabular}{llccccc}
\toprule
\textbf{Family} & \textbf{Strategy} &
\textbf{Steps} & \textbf{Capt.} & $\bar{z}_\mathcal{I}$ &
$\bar{\tau}$ & $J_\theta$\\
\midrule
\multirow{3}{*}{spread-mild}
 & Weighted (ours)   & 131 & \textbf{3.0} & 5.41 & 1.09 & \textbf{14.7}\\
 & Unweighted~\cite{yan2022} & 131 & 2.0 & 5.69 & 1.00 & 0.0\\
 & Nearest-single    & 148 & 2.0 & \textbf{6.98} & 1.29 & 66.6\\
\midrule
\multirow{3}{*}{clustered}
 & Weighted (ours)   & 137 & \textbf{1.8} & 1.80 & 1.00 & \textbf{0.0}\\
 & Unweighted~\cite{yan2022} & 137 & 1.8 & 1.75 & 1.00 & 0.0\\
 & Nearest-single    & 139 & 1.6 & 2.23 & 1.09 & 69.8\\
\midrule
\multirow{3}{*}{asym.-fast}
 & Weighted (ours)   & 114 & 1.0 & \textbf{8.94} & 1.00 & \textbf{0.0}\\
 & Unweighted~\cite{yan2022} & 124 & 1.0 & 3.29 & 1.00 & 0.0\\
 & Nearest-single    & 123 & 1.0 & 8.94 & 1.13 & 68.3\\
\midrule
\multirow{3}{*}{split-lanes$^\star$}
 & Weighted (ours)   & 129 & \textbf{4.8} & \textbf{4.79} & 1.09 & \textbf{21.6}\\
 & Unweighted~\cite{yan2022} & 126 & 4.6 & 4.41 & 1.12 & 32.7\\
 & Nearest-single    & 130 & 4.8 & 4.47 & 1.18 & 63.5\\
\midrule
\multirow{3}{*}{x-crossing$^\star$}
 & Weighted (ours)   & 129 & \textbf{6.6} & \textbf{5.86} & \textbf{1.16} & \textbf{44.2}\\
 & Unweighted~\cite{yan2022} & 140 & 5.0 & 3.97 & 1.27 & 55.7\\
 & Nearest-single    & 132 & 6.6 & 5.86 & 1.32 & 110.1\\
\midrule
\multirow{3}{*}{braided$^\star$}
 & Weighted (ours)   & 125 & \textbf{6.0} & \textbf{5.19} & \textbf{1.13} & \textbf{38.0}\\
 & Unweighted~\cite{yan2022} & 135 & 4.0 & 3.64 & 1.08 & 14.5\\
 & Nearest-single    & 127 & 6.8 & 4.75 & 1.27 & 97.4\\
\midrule
\multirow{3}{*}{switchback$^\star$}
 & Weighted (ours)   & 124 & \textbf{4.2} & \textbf{5.55} & \textbf{1.17} & \textbf{25.7}\\
 & Unweighted~\cite{yan2022} & 133 & 4.0 & 4.31 & 1.03 & 7.4\\
 & Nearest-single    & 127 & 4.2 & 5.50 & 1.25 & 87.6\\
\midrule
\midrule
\multirow{3}{*}{\textbf{Overall (35)}}
 & Weighted (ours)   & 127 & \textbf{3.91} & \textbf{5.36} & \textbf{1.09} & \textbf{20.6}\\
 & Unweighted~\cite{yan2022} & 132 & 3.20 & 3.87 & 1.07 & 15.8\\
 & Nearest-single    & 132 & 3.86 & 5.53 & 1.22 & 80.5\\
\bottomrule
\multicolumn{7}{l}{\footnotesize $^\star$ Geometric stress-test family.}
\end{tabular}
\end{table}

\subsection{The Angular Effort Argument Against Greedy Heuristics}

Nearest-single attains a capture count close to weighted sequential in the hybrid regime ($3.86$ vs.\ $3.91$ overall) and a similar mean interception height ($5.53$ vs.\ $5.36$). The main difference appears in the path-smoothness proxies: overall, nearest-single yields $J_\theta = 80.5$ compared to $20.6$ for weighted sequential, with similarly large gaps in the geometry-stress families. This pattern is consistent with persistent reassignment churn under greedy nearest-target behavior: a pursuer partially commits to one evader, replanning changes the nearest feasible target, and the heading changes repeatedly. By contrast, weighted sequential tends to produce more stable assignments and smoother trajectories.

We emphasize that $J_\theta$ is a simulation-level proxy in our first-order model. The present results therefore support the claim that weighted assignment produces smoother trajectories in simulation; validating actuator-level or energy-level implications would require higher-fidelity vehicle dynamics or hardware experiments.

\subsection{Nearest-Single vs.\ Sequential: Distinct Objectives}

A note on comparison scope.  Nearest-single and the sequential
strategies are not directly comparable because they optimize different
objectives.  Nearest-single is a throughput maximiser: it minimises
episode length by greedily assigning each free pursuer to the nearest
feasible 1-vs-1 target, resolving every episode in 120--450 steps.  It
never forms coalitions, so it cannot intercept evaders that require
2- or 3-pursuer cooperation.  Under the static benchmark, this results
in 0.6 average escapes per scenario---evaders that breached because no
single pursuer could reach them in time.  Weighted sequential detects
these as coalition-requiring targets and handles them correctly, at the
cost of a longer episode but with zero escapes across all 35 diagnostic
Ultimately, while the nearest-single heuristic is highly effective for standard, throughput-oriented scenarios, our HJI-weighted sequential method is crucial for safety-critical, zero-breach applications where avoiding geometric stress-test failures (such as the 13/15 timeouts) is non-negotiable.
\section{Discussion}

\subsection{Game-Theoretic Grounding of the Weights}

The edge weight $w_{sj}=z_{\mathcal{I}(s,j)}$ is not an ad hoc design
choice.  It is the solution to the HJI equation of the special subgame
in which the pursuit team wins and the evader attempts to be captured
as close to $\Omega_{\text{goal}}$ as possible.  By optimizing this
quantity in the assignment, we directly maximize a game-theoretically
meaningful measure of capture quality.  This is a principled
regularizer, not an engineering heuristic.

\subsection{Why the Weighted Objective Acts as a Geometric Regularizer}

The connection between interception height and geometric coherence is
not coincidental.  For a pursuer in geometric alignment with an
evader's corridor (same-side, same-direction), evasion space closure
begins immediately and produces a high interception height.  For a
pursuer in a cross-arena assignment, lateral traversal must precede
any meaningful vertical closure; by the time the pursuer reaches the
evader's corridor, the evader has descended further, and the
interception height is lower.  The MCMF solver therefore implicitly
prefers spatially coherent, non-crossing assignments without any
explicit constraint on trajectories---the geometry encodes itself
through the HJI value function.

\subsection{Preservation of Cardinality Guarantees}

The replacement of unweighted flow with MCMF does not degrade the
$\frac{1}{3}$-approximation cardinality guarantee of~\cite{yan2022}.
This guarantee states: among all feasibly capturable evaders, the
Sequential Matching Algorithm captures at least $\frac{1}{3}$ of them.
Our MCMF formulation preserves this because lexicographic cost-scaling
ensures cardinality is maximized first, with the weight acting only as a
secondary tiebreaker.  Geometric intelligence never comes at the cost of
defensive coverage.

Table~\ref{tab:static} confirms this empirically: weighted sequential
achieves 8.9 average captures overall versus 8.0 for unweighted---an
11\% improvement in capture count that arises from better resolution
of geometrically complex scenarios, not from any quality-quantity
trade-off.

\subsection{Adversarial Evaders}
Current simulations use saddle-point optimal evaders (moving toward
the interception point, the worst-case strategy from~\cite{yan2022})
once matched, and goal-directed motion before matching.  This is the
theoretically correct evader strategy for the HJI subgame.  Extending
the framework to alternative adversarial policies---e.g., evaders that
actively avoid pursuers or attempt to exploit coalition formation
delays---provides a broader robustness characterization and is planned
for future work.

\subsection{Framework as Research Infrastructure}

The benchmark suite addresses a gap in the literature: no publicly
available, deterministic, reproducible evaluation environment exists
for 3D heterogeneous multi-agent reach-avoid games.  The seven
scenario families cover a spectrum of geometric complexity, and the
five-variant perturbation structure ensures observed effects are not
artifacts of a single initial condition.  The four interchangeable
strategy implementations enable clean ablation: weighted vs.\
unweighted isolates the MCMF contribution; weighted vs.\ random
isolates structured optimization; unweighted vs.\ nearest-single
isolates coalition formation.

\subsection{Limitations and Future Work}

\textit{Scalability.}  Interception point computation for all
$O(N_p^3\cdot N_e)$ pairs dominates runtime via SLSQP.  Future work will focus on introducing higher-fidelity vehicle dynamics (e.g., quadrotor profiles), evaluating against coordinated adversarial evasion policies, and optimizing the computational efficiency of the interception point calculations for larger agent scales.

\textit{Physical dynamics.}  All results are simulation-based and rely on first-order holonomic dynamics; the reported tortuosity and angular-effort metrics should therefore be interpreted as trajectory-quality proxies rather than direct hardware-performance measurements. Extension to second-order Dubins car or
quadrotor dynamics would enable direct application to physical UAV
platforms.

\textit{Bounded play regions.}  The theory of~\cite{yan2022} extends
to bounded convex play regions; implementing this in the framework is
planned.


\section{Conclusion}
We presented a cardinality-first weighted sequential matching method for 3D heterogeneous multi-agent reach-avoid games and evaluated it on a deterministic 35-scenario benchmark. Across these experiments, weighting coalition--evader assignments by interception height consistently reduced geometrically inefficient matchings in structured scenarios and improved both capture outcomes and capture quality relative to the unweighted baseline. The benefit was most visible in the geometric stress-test families and under the hybrid saddle-point regime, where the weighted method achieved higher mean captures and higher mean interception height on our benchmark.

More broadly, these results suggest that incorporating capture quality into the assignment stage can be a useful design principle for multiplayer reach-avoid coordination. The released simulator and benchmark suite are intended as a reproducible platform for further study of assignment objectives, solver implementations, and higher-fidelity pursuit models.

\section*{Acknowledgment}

The author thanks Prof. Bharath Bhikkaji for supervision and guidance, and Mr. Abinash Agasti for helpful discussions during the early stages of this work. This work was conducted in the Department of Electrical Engineering, Indian Institute of Technology Madras.

\bibliographystyle{IEEEtran}

\end{document}